\documentclass[letterpaper, 10 pt, journal, twoside]{./IEEEtran}
\usepackage{enumitem}
\usepackage{paralist}
\usepackage{booktabs}  
\usepackage{xcolor}  
\usepackage{pifont}  
\usepackage[table]{xcolor}  

\newif\ifhighlight
\highlightfalse 
\newcommand{\highlight}[1]{
  \ifhighlight
    \textcolor{red}{#1}
  \else
    #1
  \fi
}

\highlightfalse 

\usepackage{cite}

\ifCLASSINFOpdf
  \usepackage[pdftex]{graphicx}
\else
\fi
\usepackage{amsmath}
\usepackage{amsfonts,amssymb}
\usepackage{algorithm}    
\usepackage{algorithmic}

\usepackage{array}
\begin{document}
%
\title{MonoSE(3)-Diffusion: A Monocular SE(3) Diffusion Framework for Robust Camera-to-Robot Pose Estimation}
%
%
%

\author{Kangjian Zhu$^{1}$, Haobo Jiang$^{2}$, Yigong Zhang$^{3*}$, Jianjun Qian$^{1}$, Jian Yang$^{1}$, Jin Xie$^{4*}$%
\thanks{Manuscript received: April, 19, 2025; Revised June, 16, 2025; Accepted August, 29, 2025.}
\thanks{This paper was recommended for publication by Markus Vincze upon evaluation of the Associate Editor and Reviewers' comments.
This work was supported by the National Key R\&D Program of China No. 2024YFC3015801, the National Science and Technology Major Project No. 2022ZD0116305, and National Science Fund of China under Grant Nos. 62361166670, U24A20330, 62276144 and 62306155.
} 
\thanks{$^{1}$Kangjian Zhu, Jianjun Qian and Jian Yang are with School of Computer Science and Engineering, Nanjing University of Science and Technology, China,
        {\tt\footnotesize zkangjian@njust.edu.cn}}%
\thanks{$^{2} $Haobo Jiang is with ANGEL CorpLab and College of Computing and Data Science, Nanyang Technological University, Singapore,
        {\tt\footnotesize haobo.jiang@ntu.edu.sg}}%
\thanks{$^{3}$Yigong Zhang is with College of Computer Science, Nankai University, China,
        {\tt\footnotesize zyg025@nankai.edu.cn}}%
\thanks{$^{4} $Jin Xie is with School of Intelligence Science and Technology, Nanjing University, China,
        {\tt\footnotesize csjxie@nju.edu.cn}}%
\thanks{$^{*}$Corresponding authors: Yigong Zhang and Jin Xie.}
\thanks{Digital Object Identifier (DOI): see top of this page.}
}
%
%

\markboth{IEEE Robotics and Automation Letters. Preprint Version. Accepted September, 2025}
{Zhu \MakeLowercase{\textit{et al.}}: MonoSE(3)-Diffusion} 

\maketitle

\begin{abstract}
We propose MonoSE(3)-Diffusion, a monocular SE(3) diffusion framework that formulates markerless, image-based robot pose estimation as a conditional denoising diffusion process.
The framework consists of two processes: a visibility-constrained diffusion process for diverse pose augmentation and a timestep-aware reverse process for progressive pose refinement.
The diffusion process progressively perturbs ground-truth poses to noisy transformations for training a pose denoising network.
Importantly, we integrate visibility constraints into the process, ensuring the transformations remain within the camera field of view.
Compared to the fixed-scale perturbations used in current methods, the diffusion process generates in-view and diverse training poses, thereby improving the network generalization capability.
Furthermore, the reverse process iteratively predicts the poses by the denoising network and refines pose estimates by sampling from the diffusion posterior of current timestep, following a scheduled coarse-to-fine procedure. 
%
Moreover, the timestep indicates the transformation scales, which guide the denoising network to achieve more accurate pose predictions.
The reverse process demonstrates higher robustness than direct prediction, benefiting from its timestep-aware refinement scheme.
Our approach demonstrates improvements across two benchmarks (DREAM and RoboKeyGen), achieving a notable AUC of 66.75 on the most challenging dataset, representing a 32.3\% gain over the \textit{state-of-the-art}.
\end{abstract}


%
\IEEEpeerreviewmaketitle

\section{Introduction}
%
%
%
%
\IEEEPARstart{R}{ecovering} the 6-DoF camera-to-robot pose from a monocular image is a fundamental task with numerous applications, including robotic grasping \cite{graspnet}, manipulation \cite{demonstrate_once}, and rearrangement \cite{structdiffusion2023}. 
While traditional offline methods, such as hand-eye calibration \cite{Hand-Eye-Calibration-CVPR}, have long been standard, the advent of deep learning has introduced markerless pose estimation techniques that are practical, cost-effective, and highly effective for dynamic and unstructured environments.

Mainstream deep learning-based camera-to-robot pose estimation can be broadly categorized into two types: keypoint-based methods \cite{DREAM, Temporal-keypoint-2023cvpr, Self-supervised-robot-pose} and regression-based methods \cite{holistic, robopose}.
Keypoint-based methods rely on the Perspective-n-Point (PnP) process, which is highly sensitive to inaccuracies in keypoint prediction, leading to poor performance when the robot is obscured or unclear.
In contrast, regression-based methods leverage the robot's full geometric structure to predict transformations and iteratively refine poses, yielding higher accuracy.
%
However, these methods still exhibit limited precision when handling out-of-distribution poses or low-visibility robot targets.
We identify two key limitations in current approaches:
(\textbf{i}) Training poses are generated through fixed-scale perturbations of ground-truth poses, limiting diversity and compromising network generalization.
(\textbf{ii}) The estimation process relies solely on network predictions, lacking a scheduled refinement pipeline and robustness against inaccuracies.

\begin{figure}[t]  \centering
   \includegraphics[width=\linewidth]{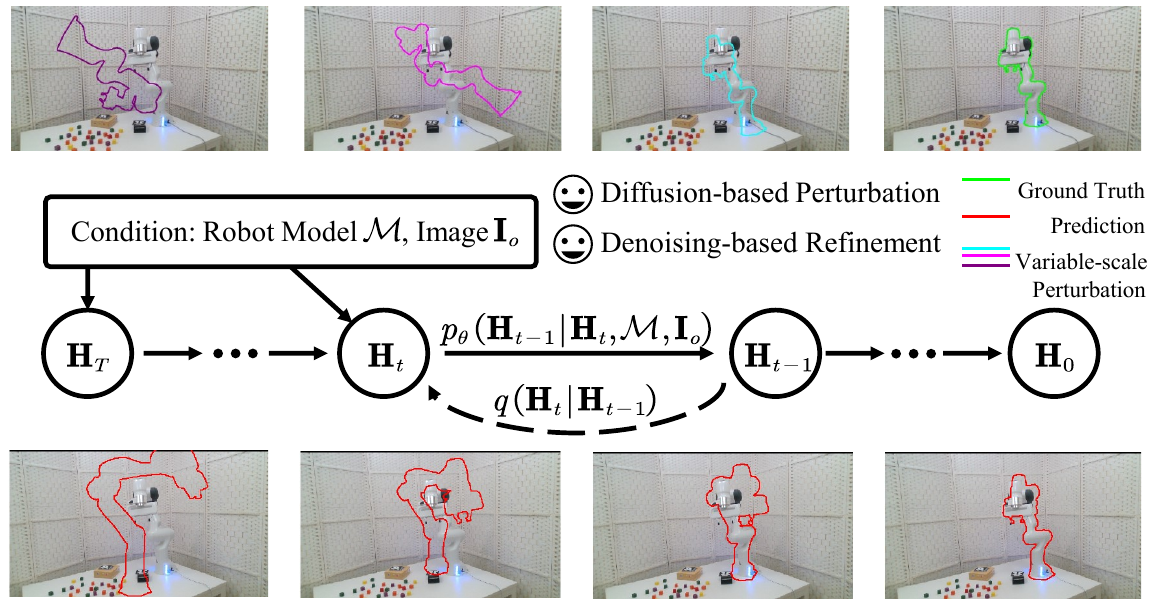}
   \vspace{0.2cm} 
   {\footnotesize (a) Pose estimation based on the conditional denoising diffusion process.}
   \includegraphics[width=\linewidth]{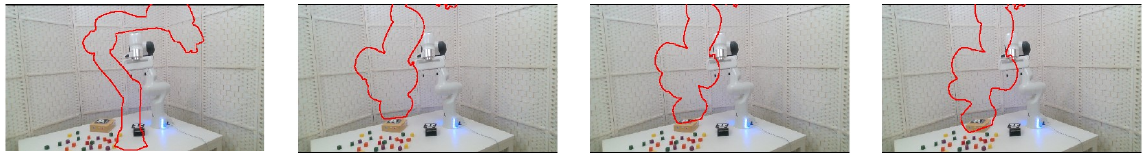}
   {\footnotesize (b) Iterative pose regression.}
   \caption{
   Comparison of camera-to-robot pose estimation process between our proposed diffusion-based method and regression-based method.
   (\textbf{a}) MonoSE(3)-Diffusion leverages a diffusion process to generate diverse training poses, and a scheduled reverse diffusion process to achieve coarse-to-fine pose estimation, conditioned on the robot model and the input image.
   (\textbf{b}) Conventional iterative regression suffers from inaccurate prediction and premature convergence.
   \textbf{Note}: The poses are visualized using edges, representing the mask boundaries of the rendered view under the poses.
   }
   \vspace{-0.5cm}
   \label{fig:introduction}
\end{figure}

In this paper, we propose a \textbf{Mono}cular \textbf{SE(3)} \textbf{Diffusion} framework (\textbf{MonoSE(3)-Diffusion}), a diffusion model-based pose estimation method for a monocular image-based input.
We formulate the camera-to-robot pose estimation as a denoising diffusion process conditioned on a robot model and image, as shown in Fig.~\ref{fig:introduction}(a).
%
Our framework consists of two core processes: a visibility-constrained SE(3) diffusion process and a timestep-aware SE(3) reverse process, which address the two limitations mentioned above, respectively.
In the training phase, the diffusion process progressively samples noisy transformations to perturb ground-truth poses, generating diverse pose samples to train a pose denoising network.
To ensure the pose samples to satisfy the requirements of monocular tasks, we integrate the constraints of the camera field of view into the transformations.
%
%
%
Thus, these pose samples exhibit both diversity and proper distribution within viewing frustum, enabling comprehensive network training and effectively addressing limitation (\textbf{i}).
%
In the inference phase, the reverse process iteratively refines pose estimates to align with the conditional input images, based on the trained denoiser.
%
Notably, at each pose refinement step, the estimation is guided by the reverse process posterior, which dynamically balances coarse exploration and fine-grained refinement across timesteps.
%
%
%
Furthermore, we modify a rendering-based pose regression model~\cite{robopose} as the denoising network, incorporating timestep as an additional conditioning input to facilitate accurate predictions.
Compared to direct iterative regression (Fig.~\ref{fig:introduction}(b)), our timestep-aware estimation mitigates premature convergence risks and further enhances the prediction robustness, thereby addressing limitation (\textbf{ii}).
%

Our main contributions are summarized as follows:
\begin{compactitem}
     \item 
    We propose a monocular SE(3) diffusion framework, which formulates the image-based pose refinement as the conditional denoising diffusion process in the viewing frustum for robust camera-to-robot pose estimation.
    \item 
         We introduce a visibility-constrained diffusion process to ensure the generated poses maintain both in-view distribution and diversity for comprehensive model training, significantly enhancing generalization capability.
    \item 
        We introduce a timestep-aware reverse process to progressively refine poses by the diffusion posterior of each step, 
        following a designed coarse-to-fine estimation procedure to achieve robust results.
\end{compactitem}

%

\section{Related Work}
\label{sec:related_work}
%
Hand-eye calibration~\cite{Hand-Eye-Calibration-CVPR} is a traditional method for camera-to-robot pose estimation, typically using fiducial markers such as AprilTag.
Recent research has increasingly adopted markerless, deep learning-based approaches owing to their demonstrated practicality, cost-effectiveness, and superior performance in dynamic, unstructured environments.
DREAM~\cite{DREAM} provides synthetic training and real-world test data, laying the foundation for learning-based pose estimation.
RoboKeyGen~\cite{robokeygen} introduces a more challenging real-world dataset, where robots are visually indistinct from the background or appear in small regions, posing difficulties for existing methods.
Mainstream methods are broadly categorized as \textbf{keypoint-based} or \textbf{regression-based}, following common practice in object pose estimation~\cite{DeepIM,jiang2023se3}.

\subsection{Keypoint-based pose estimation}

These methods detect 2D keypoints and estimate the camera-to-robot pose via 2D–3D correspondence with PnP, based on the forward  kinematic model parameterized by known joint angles.
DREAM~\cite{DREAM} employs a stacked hourglass network for 2D keypoint detection.
%
SGTAPose~\cite{Temporal-keypoint-2023cvpr} leverages robot structural priors and temporal continuity to resolve single-view occlusions and ambiguities.
%
CtRNet~\cite{Self-supervised-robot-pose} bridges the sim-to-real gap by linking pose estimation to robot mask differences for self-supervision.
CtRNet-X~\cite{ctrnet-x} further integrates vision-language models into the network to enable fine-grained detection of robot components, addressing the issue of invisible parts.
RoboKeyGen~\cite{robokeygen} extends SGTAPose by using a diffusion model to lift 2D keypoints detection to 3D space without relying on image features. 
Unlike prior methods that require known joint angles, it predicts joint angles directly from generated 3D keypoints.
While keypoint-based methods are effective, their accuracy is inherently limited due to the sensitivity to errors in keypoint predictions.
%

\subsection{Regression-based pose estimation}

In contrast to keypoint-based methods, these approaches infer 3D information directly from images, including the robot's pose and joint angles.
%
RoboPose~\cite{robopose} adopts a \textit{render\&compare} paradigm~\cite{DeepIM} to iteratively refine pose estimates, based on the robot mesh.
%
HoliPose~\cite{holistic} proposes a decoupled framework for rotation, depth, and joint angle regression, but requires the robot's bounding box to be known in the image.
RoboPEPP~\cite{robopepp} enhances robot structure understanding through a self-supervised embedding-predictive framework, with robot bounding boxes obtained via a vision-language model.
While regression-based methods surpass keypoint approaches in accuracy, their performance degrades significantly under challenging conditions, as evidenced by RoboKeyGen~\cite{robokeygen} benchmark results.
%
%
To address this issue, we propose a diffusion framework for robust camera-to-robot pose estimation.

\subsection{SE(3) pose diffusion framework}

Inspired by the success of denoising diffusion probabilistic models (DDPM)~\cite{DDPM}, SE(3) diffusion model have been increasingly applied to point cloud registration \cite{wu2024diff, jiang2023se3, jiang2021sampling, jiang2023robust}.
%
In robotic applications, 
SE(3)-DiffusionFields \cite{se3diffusionfields} leverages diffusion models to learn a grasping pose cost for 6DoF grasping.
StructDiffusion \cite{structdiffusion2023} employs a diffusion model to predict the relative transformation poses of the objects for assembly and rearrangement.
%
%
However, current SE(3) diffusion models struggle with monocular inputs due to unconstrained out-of-view pose generation.
In this paper, we propose a diffusion process 
which maintains pose diversity while ensuring that the sampled poses remain within the camera's field of view.

\begin{figure*}[t]
  \centering
   \includegraphics[width=\linewidth]{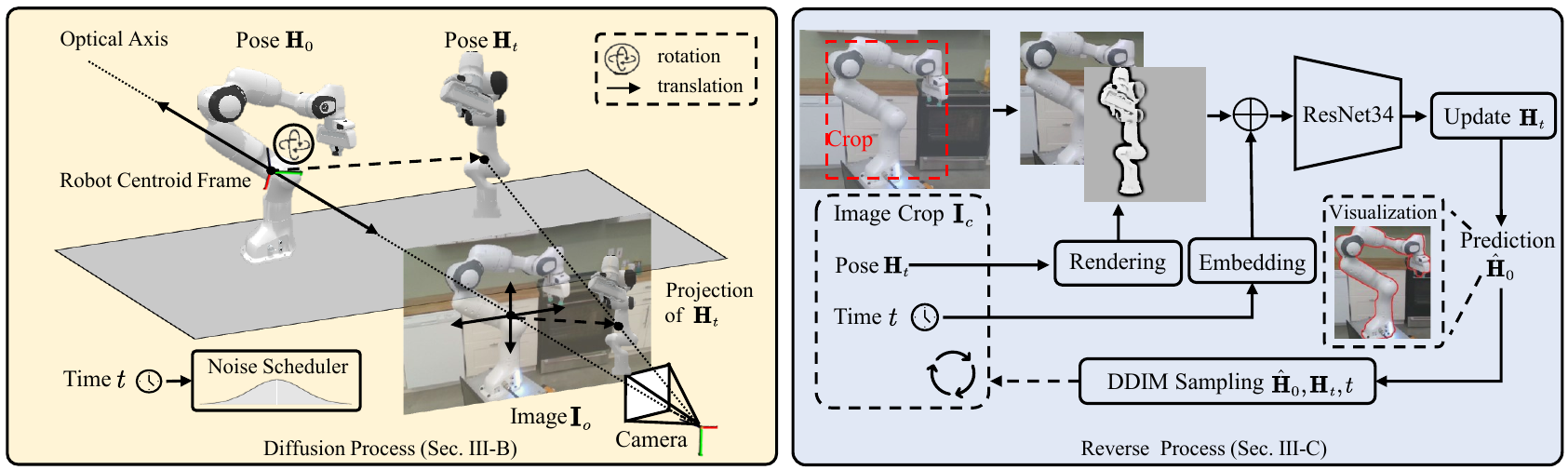}
   \caption{
   \textbf{Diffusion process.} The visibility-constrained diffusion process is specifically designed to satisfy the viewing frustum constraints, and is decoupled into centroid-based rotation, translation in the image plane, and translation along the optical axis.
   It takes the ground-truth pose $\mathbf{H}_0$ and the image as inputs, generating a noisy pose $\mathbf{H}_t$ along with a sampled time step $t$. 
   The projections of noisy poses are bounded by the camera viewing frustum.
   \textbf{Reverse process.} The timestep-aware reverse process is driven by DDIM sampling and a pose denoising network.
   The network represents the noisy pose as a rendered view, which is concatenated with the cropped image and the embedded time step before being processed by a ResNet.
   Subsequently, a pose update strategy is employed to get the pose prediction $\hat{\mathbf{H}}_0$, which is obtained by DDIM sampling to generate a progressively denoised pose $\mathbf{H}_{t-1}$ for the next iteration. 
   }
   \label{fig:method}
   \vspace{-0.5cm}
\end{figure*}

\section{Method}
\label{sec:method}
In this section, we present MonoSE(3)-Diffusion, a diffusion framework for robust camera-to-robot pose estimation from a monocular image, as shown in Fig.~\ref{fig:method}.
The task definition and the preliminaries of the diffusion framework are outlined in Sec.~\ref{sec:method-1}. 
%
We introduce two key components in our framework: the visibility-constrained diffusion process (VisDiff) and the timestep-aware reverse process (RevDiff). 
These are detailed in Sec.~\ref{sec:method-2} and Sec.~\ref{sec:method-3}, respectively.
The training and inference processes are detailed in Alg.~\ref{train_algorithm} and Alg.~\ref{infer_algorithm}, respectively.

\begin{figure}[t]
  \centering
   \includegraphics[width=\linewidth]{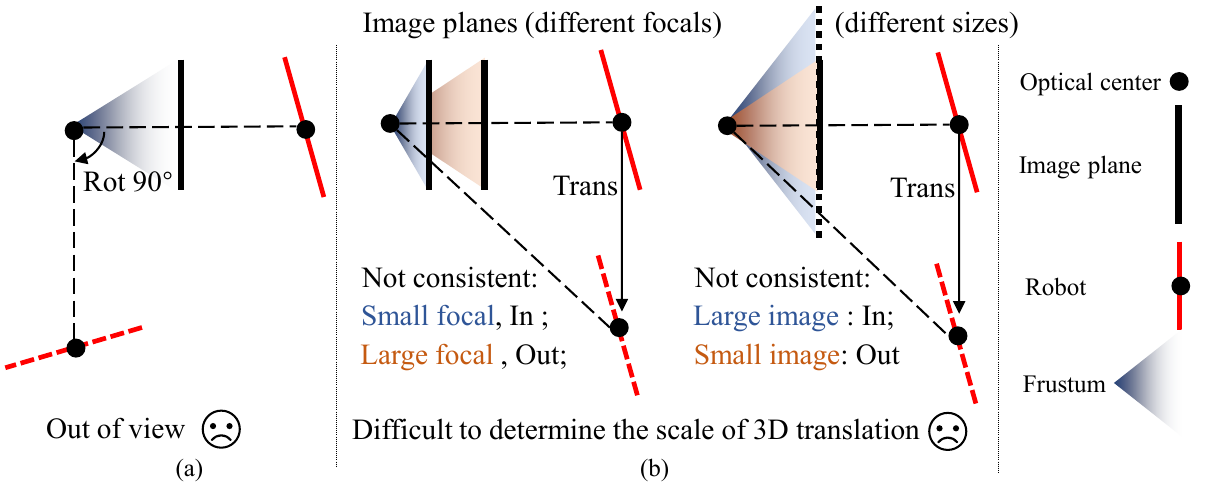}
   \caption{
   Visibility constraints in monocular images. 
   (\textbf{a}) Rotation around the camera's optical center (defined in the camera coordinate frame) can easily transform the original pose out of the camera's field of view.
   (\textbf{b}) The same translation can lead to different shifts in the image plane, depending on focal length (left) and image size (right), making it difficult to define consistent diffusion parameters for 3D translation.  
   }
   \label{fig:geometry}
   \vspace{-0.4cm}
\end{figure}

\subsection{Background}
\label{sec:method-1}
\textbf{Task definition.} 
Our objective is to determine the robot pose $\mathbf{H}\in SE(3)$ in the camera coordinate frame, which consists of orientation $\mathbf{R} \in SO(3)$ and position $\mathbf{t}\in \mathbb{R}^3$.
%
The ground-truth pose $\mathbf{H}_0$ (subscript denoting the diffusion timestep), is represented by a $4\times4$ homogeneous transformation matrix as follows:
\begin{equation}
    \mathbf{H}_0 = \begin{bmatrix} \mathbf{R}_{0} & \mathbf{t}_{0} \\ \mathbf{0}^\top & 1 \end{bmatrix}
\end{equation}
which we denote shorthand as $\mathbf{H}_0 = (\mathbf{R}_{0}, \mathbf{t}_{0})$.
The conditions include a monocular original image $\mathbf{I}_o\in \mathbb{R}^{w\times h\times3}$, robot joint angle configurations $\mathcal{J}$ and a robot 3D model.
Notably, the model varies with $\mathcal{J}$ through robot kinematics and is denoted shorthand as $\mathcal{M}(\mathcal{J})$ in this paper.
\newline
\textbf{Preliminary of Euclidean diffusion.}  
The diffusion process aims to generate noisy data by successively injecting Gaussian noise into the data sample.
Specifically, given a sample $\mathbf{x}_0\sim p_{data}$ from the dataset, the process creates a Markov chain of latent variables $\mathbf{x}_i\mid_{i=1}^{T}$ from distribution $q(\mathbf{x}_t\mid \mathbf{x}_{t-1})\sim \mathcal{N}(\mathbf{x}_{t};\sqrt{1-\beta_t}\mathbf{x}_{t-1}, \beta_t\mathbf{I})$, where $\beta_{t}$ is the noise coefficient determined by a schedule.
To reduce computational overhead, the variable $\mathbf{x}_t$ is also available through a closed-form formulation:
\begin{equation}
    \mathbf{x}_t = \sqrt{\bar{\alpha}_t} \mathbf{x}_0 + \sqrt{1 - \bar{\alpha}_t} \, \boldsymbol{\epsilon}, \quad \boldsymbol{\epsilon} \sim \mathcal{N}(0, \mathbf{I}),
    \label{eq:Euclidean-diffusion}
\end{equation}
where the coefficients $\bar{\alpha}_t = \prod_{s=0}^t (1 - \beta_t)$.

The reverse diffusion process aims to iteratively denoise the noisy sample, returning it to the original data distribution.
This defines conditional probabilities $p_\theta(\mathbf{x}_{t-1} \mid \mathbf{x}_t) \sim \mathcal{N}(\mathbf{x}_{t-1}; \boldsymbol{\mu}_{\theta}(\mathbf{x}_t, t), \sigma_t^2 \mathbf{I})$, where $\boldsymbol{\mu}_{\theta}(\mathbf{x}_t, t)$ is the mean prediction and $\sigma_t^2$ is the variance.
The \textbf{DDIM} (Denoising Diffusion Implicit Models) sampling algorithm is a variant of the traditional denoising diffusion models, introduced to enable faster and more flexible sampling~\cite{ddim}. 
The formulation for DDIM is given as follows:
\begin{equation}
    \mathbf{x}_{t-1} = \sqrt{\bar{\alpha}_{t-1}} \, \hat{\mathbf{x}}_0 + \sqrt{1 - \bar{\alpha}_{t-1} - \sigma_t^2} \, \boldsymbol{\epsilon}_\theta(\mathbf{x}_t, t) + \sigma_{t} \, \mathbf{z},
    \label{eq:reverse_diffusion}
\end{equation}
where $\hat{\mathbf{x}}_0$ and $\boldsymbol{\epsilon}_\theta(\mathbf{x}_t, t)$ represent the prediction of sample and added noise in diffusion process Eq. \ref{eq:Euclidean-diffusion}.
Additionally, $\sigma_t = \eta \sqrt{\frac{(1 - \bar{\alpha}_{t-1} / \bar{\alpha}_t)(1 - \bar{\alpha}_t)}{1 - \bar{\alpha}_{t-1}}}$, with $\eta$ as a scaling parameter that controls the amount of added noise $\mathbf{z}\sim \mathcal{N}(0, \mathbf{I})$ in the reverse diffusion.

\subsection{Visibility-constrained SE(3) diffusion process}
\label{sec:method-2}
The objective of SE(3) diffusion is to generate widely distributed noisy pose samples by transforming the ground-truth pose $\mathbf{H}_0$, following the general diffusion process defined in Eq.~\ref{eq:Euclidean-diffusion}.
In general, existing SE(3) diffusion approaches~\cite{se3diffusionfields, jiang2023se3} sample the random transformations corresponding to the noise $\boldsymbol{\epsilon}$ in Eq.~\ref{eq:Euclidean-diffusion}.
%
However, these methods sample poses unconstrained in SE(3), violating the in-frustum prior essential for monocular tasks where the pose must project the robot within the camera's viewing frustum.
\newline
\textbf{Constraint of the viewing frustum.}
Concretely, the frustum constrains the position component of the generated pose, while the orientation component remains unrestricted and must be distributed across the SO(3) space to ensure comprehensive training.
%
The diversity requirement for orientation prevents us from naively modifying the diffusion parameters of existing methods to satisfy the position constraint, as this would severely limit the orientation diversity of the generated pose samples.
As such, we expect the monocular SE(3) diffusion to sample transformations that rotate the pose across the SO(3) space while translating it within the frustum.
To achieve this, we separately analyze the effects of the \textbf{rotation} and \textbf{translation} components in the transformation, as illustrated in Fig.~\ref{fig:geometry}.
To clarify, orientation and position are components of the pose, while rotation and translation are components of the sampled transformation for the pose diffusion process.
%
\newline
\textbf{Rotation.}
We find that rotation can unintentionally induce significant position changes, as shown in Fig.~\ref{fig:geometry}(a), potentially moving the robot out of view.   
The primary reason is that rotation is defined in the camera coordinate frame, leading to a correlation between rotation and translation.
To address this issue, we redefine the rotation in the robot-centroid coordinate frame, as illustrated in Fig.~\ref{fig:method}(a), effectively decoupling the correlation between rotation and translation.
This approach is inspired by \cite{DeepIM}, which also decouples the correlation between rotation and translation for independent regression.
We avoid using the robot's local coordinate frame, as rotation defined in this frame can still cause shifts in the robot position.
However, for general object pose estimation, the object local frame remains suitable.
\newline
\textbf{Translation.}
According to perspective projection, the camera intrinsics influence the shape of the viewing frustum, leading to different position constraints.
As shown in Fig.~\ref{fig:geometry}(b), the same translation can yield different results due to variations in focal length and image size.
This inconsistency presents a challenge in formulating a diffusion process that can generalize across variable scenes with different camera intrinsics.
To address this challenge, we decompose the translation into components along the image plane and the camera optical axis, as shown in Fig.~\ref{fig:method}.(a), which incorporates the focal length and image size into the diffusion process.
%
\newline
\textbf{Monocular diffusion process.}
%
Formally, for pose $\mathbf{H}_0 = (\mathbf{R}_{0}, \mathbf{t}_{0})$, where $\mathbf{t}_{0}=[\mathbf{t}_{0}^{x}, \mathbf{t}_{0}^{y}, \mathbf{t}_{0}^{z}]$ denotes the translation vector and $\mathbf{R}_{0} = [\mathbf{r}_{0}^{1}, \mathbf{r}_{0}^{2}, \mathbf{r}_{0}^{3}]$ is the rotation matrix with column vectors $\mathbf{r}_{0}^{1}, \mathbf{r}_{0}^{2}, \mathbf{r}_{0}^{3}$, 
we define the monocular-normalized formulation as:
\begin{equation}
    \tilde{\mathbf{H}}_0 = (\underbrace{\mathbf{r}_{0}^{1}, \mathbf{r}_{0}^{2}}_{\tilde{\mathbf{R}}_{0}}, \,\underbrace{\frac{f\mathbf{t}_{0}^{x}}{w\mathbf{t}_{0}^{z}}, \frac{f\mathbf{t}_{0}^{y}}{h\mathbf{t}_{0}^{z}}, \mathbf{t}_{0}^{z} - \mathrm{c}_z)}_{\tilde{\mathbf{t}}_{0}},
    \label{eq:monocular-normalized}
\end{equation}
where $w, h$ are the image sizes, $f$ is camera focal length, and $\mathrm{c}_z$ is a predefined depth normalization parameter.
This formulation is associated with SE(3), while also allowing Euclidean diffusion through Eq.~\ref{eq:Euclidean-diffusion}, which offers two advantages:
(\textbf{i}) 
The diffusion of $\tilde{\mathbf{R}}_{0} \in \mathbb{R}^{6}$ is applied independently of translation, ensuring that only the robot's orientation is perturbed without affecting its position.
Following StructDiffusion~\cite{structdiffusion2023}, the perturbation of the 6D rotation representation enables effective and stable sampling over SO(3) during the diffusion process.
(\textbf{ii}) The scale of $\tilde{\mathbf{t}}_{0}$ remains invariant to various camera intrinsics, enabling the generated poses to be adjusted and distributed throughout the viewing frustum.
By applying the diffusion in Eq.~\ref{eq:Euclidean-diffusion} to $\tilde{\mathbf{H}}_0$, we can obtain $ \tilde{\mathbf{H}}_t$ as follows. 
\begin{equation}
    \tilde{\mathbf{H}}_t = \sqrt{\bar{\alpha}_t} \tilde{\mathbf{H}}_0 + \sqrt{1 - \bar{\alpha}_t} \, \boldsymbol{\epsilon}, \quad \boldsymbol{\epsilon} \sim \mathcal{N}(0, \mathbf{I}).
    \label{eq:monocular-Euclidean-diffusion}
\end{equation}
$\tilde{\mathbf{H}}_t$ is the monocular-normalized formulation of the generated noisy pose ${\mathbf{H}}_t$.
To obtain the pose $\mathbf{H}_t=(\mathbf{R}_{t}, \mathbf{t}_{t})$, we apply an inverse normalization process of Eq.~\ref{eq:monocular-normalized} to the $ \tilde{\mathbf{H}}_t$.
The rotation $\mathbf{R}_{t} = \left[ \mathbf{u}_t^1 \; \mathbf{u}_t^2 \; \mathbf{u}_t^3 \right]$ is a $3 \times 3$ rotation matrix obtained by Gram-Schmidt orthogonalization as follow:
\begin{equation}
\begin{aligned}
    \hfill \mathbf{u}_t^1 &= \mathbf{r}_t^1, \quad \mathbf{u}_t^2 = \mathbf{r}_t^2 - \frac{\mathbf{r}_t^2 \cdot \mathbf{u}_t^1}{\mathbf{u}_t^1 \cdot \mathbf{u}_t^1} \mathbf{u}_t^1 \\
    \hfill \mathbf{u}_t^1 &= \frac{\mathbf{u}_t^1}{\|\mathbf{u}_t^1\|}, \quad \hfill \mathbf{u}_t^2 = \frac{\mathbf{u}_t^2}{\|\mathbf{u}_t^2\|},\quad \hfill \mathbf{u}_t^3 = \mathbf{u}_t^1 \times \mathbf{u}_t^2. \\
    \label{monocular-denormalized-rotation}
\end{aligned}
\end{equation}
The translation $\mathbf{t}_t = \left[\mathbf{t}_t^x \; \mathbf{t}_t^y\; \mathbf{t}_t^z\right]$ is obtained as follow:
\begin{equation}
\begin{aligned}
    \hfill \mathbf{t}_t^x &= \frac{w \mathbf{t}_t^z}{f} \tilde{\mathbf{t}}_t^x, \quad \hfill \mathbf{t}_t^y = \frac{h \mathbf{t}_t^z}{f} \tilde{\mathbf{t}}_t^y, \quad \hfill \mathbf{t}_t^z = \tilde{\mathbf{t}}_t^z + \mathrm{c}_z.
    \label{monocular-denormalized-translation}
\end{aligned}
\end{equation}
As such, we define the monocular pose diffusion $q(\mathbf{H}_t\mid\mathbf{H}_{0})$ with a monocular-normalized formulation of SE(3) introduced.
Using Eqs.~4-7, the diffusion process ensures sample diversity while preserving view frustum constraints.
To clarify, the $\mathbf{H}$ represents the $4 \times 4$ homogeneous pose matrix; $\tilde{\mathbf{H}}$ is the monocular-normalized formulation, and $\hat{\mathbf{H}}$ is the $4 \times 4$ pose prediction defined in Sec.~\ref{sec:method-3}. 
\begin{algorithm}[tp]
    \small
    \caption{Training Phase}
    \begin{algorithmic}[]  
        \STATE \textbf{Initialize} the pose denoising network $\mathcal{D}_{\theta}$.
        \WHILE{$\mathcal{D}_{\theta}$ has not converged}
            \STATE \textbf{Sample} an image $\mathbf{I}_o$, ground-truth pose $\mathbf{H}_0$, and joint-conditional robot model $\mathcal{M}(\mathcal{J})$ from the dataset.
            \STATE \textbf{Sample} a diffusion time step $t \sim [1, T]$.
            \STATE \textbf{Generate} the noisy pose $\mathbf{H}_t$ using the visibility-constrained diffusion process:
            \STATE \quad \quad \quad $q\left( \mathbf{H}_t \right. |\left. \mathbf{H}_0 \right)$ \textbf{following} Eqs.~4-7.
            \STATE \textbf{Predict} the denoised pose using the pose denoising network:
            \STATE \quad \quad \quad $\hat{\mathbf{H}}_0 = \mathcal{D}_{\theta}(\mathbf{H}_t | \mathbf{I}_o, t, \mathcal{M}(\mathcal{J}))$ \textbf{following} Eq.~\ref{eq:pose_denoising}.
            \STATE \textbf{Compute} the loss $\mathcal{L}_\theta$ \textbf{following} Eq.~\ref{eq:full_supervised}.
            \STATE \textbf{Update} the pose denoising network $\mathcal{D}_{\theta}$.
        \ENDWHILE
    \end{algorithmic}
    \label{train_algorithm}
\end{algorithm}
\begin{algorithm}[tp]
    \small
    \caption{Inference Phase}
    \begin{algorithmic}[]  
        \STATE \textbf{Input}: image $\mathbf{I}_o$, joint-conditional robot model $\mathcal{M}(\mathcal{J})$, pose denoising network $\mathcal{D}_{\theta}$.
        \STATE \textbf{Initialize} pose estimation $\mathbf{H}_T$.
        \FOR{$t = T$ to $1$}
            \STATE \textbf{Sample} pose $\mathbf{H}_{t-1}$ from the distribution of reverse diffusion process $p_\theta(\mathbf{H}_{t-1} \mid \mathbf{H}_t, \mathcal{M}(\mathcal{J}), \mathbf{I}_o)$:
            \STATE \quad \quad \textbf{Obtain} $\tilde{\mathbf{H}}_t$ from $\mathbf{H}_t$ \textbf{following} Eq.~\ref{eq:monocular-normalized}.
            \STATE \quad \quad \textbf{Predict} the denoised pose:
            \STATE \quad \quad \quad \quad \quad $\hat{\mathbf{H}}_0 = \mathcal{D}_{\theta}(\mathbf{H}_t | \mathbf{I}_o, t, \mathcal{M}(\mathcal{J}))$ \textbf{following} Eq.~\ref{eq:pose_denoising}.
            \STATE \quad \quad \textbf{Obtain} $\tilde{\mathbf{H}}_0$ from $\hat{\mathbf{H}}_0$ \textbf{following} Eq.~\ref{eq:monocular-normalized}.
            \STATE \quad \quad \textbf{Generate} $\tilde{\mathbf{H}}_{t-1}$ using DDIM sampling conditioned on $\tilde{\mathbf{H}}_0$ and $\tilde{\mathbf{H}}_t$ \textbf{following} Eq.~\ref{eq:monocular-Euclidean-reverse}.
            \STATE \quad \quad \textbf{Obtain} $\mathbf{H}_{t-1}$ from $\tilde{\mathbf{H}}_{t-1}$ \textbf{following} Eq.~\ref{monocular-denormalized-rotation} and Eq.~\ref{monocular-denormalized-translation}.
        \ENDFOR
    \end{algorithmic}
    \label{infer_algorithm}
\end{algorithm}

\subsection{Timestep-aware reverse process}
\label{sec:method-3}
In this section, the generated noisy pose $\mathbf{H}_t$ is denoised by sampling from the reverse diffusion posterior, where the timestep is taken into account to avoid premature convergence. 
The denoising process is carried out by a pose denoising network $\mathcal{D}_{\theta}$, which predicts the camera-to-robot pose $\hat{\mathbf{H}}_0$ conditioned on the original input image $\mathbf{I}_o$.
We begin by presenting the formulation of reverse diffusion, following the sampling process defined in Eq.~\ref{eq:reverse_diffusion}.
Then, we introduce the rendering-based pose denoising network in detail, along with its supervised loss.
\newline
\textbf{Formulation.}
Conditioned on the generated pose $\mathbf{H}_t$ and the predicted pose $\hat{\mathbf{H}}_0$, we expect to iteratively denoise the pose to $\mathbf{H}_{t-1}$.
Note that the three poses are all $4\times4$ homogeneous pose matrices, but their monocular-normalized formulations are compatible with the DDIM sampling in Eq.~\ref{eq:reverse_diffusion} as follows:
\begin{equation}
\begin{aligned}
    \tilde{\mathbf{H}}_{t-1} = \sqrt{\bar{\alpha}_{t-1}} \, \tilde{\mathbf{H}}_0 + \sqrt{1 - \bar{\alpha}_{t-1} - \sigma_t^2} \, \boldsymbol{\epsilon}_\theta , 
\end{aligned}
\label{eq:monocular-Euclidean-reverse}
\end{equation}
where $\boldsymbol{\epsilon}_\theta=\frac{\tilde{\mathbf{H}}_t-\sqrt{\bar{\alpha}_t} \tilde{\mathbf{H}}_0}{\sqrt{1 - \bar{\alpha}_t}}$.
Notably, $\tilde{\mathbf{H}}_0$, $\tilde{\mathbf{H}}_{t-1}$ and $\tilde{\mathbf{H}}_{t}$ represent the monocular-normalization of $\hat{\mathbf{H}}_0$, $\mathbf{H}_{t-1}$ and $\mathbf{H}_{t}$, and $\boldsymbol{\epsilon}_\theta$ is the predicted noise added in Eq.~\ref{eq:monocular-Euclidean-diffusion}.
Additionally, the stochastic noise term $\sigma_t \mathbf{z}$ in Eq.~\ref{eq:reverse_diffusion} is omitted, which is a common practice in diffusion models for fine-grained tasks such as pose estimation \cite{jiang2023se3}.
In summary, we can sample the pose from the posterior distribution of reverse diffusion process $p_\theta(\mathbf{H}_{t-1} \mid \mathbf{H}_t)$, by first obtaining the monocular-normalized pose (Eq.~\ref{eq:monocular-normalized}), then performing DDIM sampling (Eq.~\ref{eq:monocular-Euclidean-reverse}), and finally remapping to $4\times4$ pose (Eq.~\ref{monocular-denormalized-rotation} and Eq.~\ref{monocular-denormalized-translation}), with the procedure detailed in Alg.~\ref{infer_algorithm}.
\newline
\textbf{Pose denoising network.}
The pose denoising network takes the generated pose $\mathbf{H}_t=(\mathbf{R}_{t},\mathbf{t}_{t})$ as input and outputs the pose prediction $\hat{\mathbf{H}}_{0}=(\hat{\mathbf{R}}_{0},\hat{\mathbf{t}}_{0})$, conditioned on the original image $\mathbf{I}_o\in \mathbb{R}^{w\times h\times3}$, robot joint angle configurations $\mathcal{J}$, and diffusion step $t$.
We obtain the robot 3D model in current joint configurations as $\mathcal{M}(\mathcal{J})$, and generate its imagined view in pose $\mathbf{H}_t$ using a standard renderer, which gives us a rendered image $\mathbf{I}_{r}\in \mathbb{R}^{w_{r}\times h_{r}\times3}$, where $w_{r}$ and $h_{r}$ are hyperparameters that determine the size of rendered image.
Then we expect a crop of the input image $\mathbf{I}_{c}$ with the same size as the rendered image, which is obtained by projecting the model $\mathcal{M}(\mathcal{J})$ from $\mathbf{H}_t$ onto the original image $\mathbf{I}_o$, cropping the corresponding region, and scaling it to the target size.
%
The detailed process is qualitatively demonstrated in the Appendix.
\newline
\textbf{Pose prediction.}
The network is designed to compare the rendered image $\mathbf{I}_{r}$ with the cropped input image $\mathbf{I}_{c}$.
The diffusion step $t \in \mathbb{R}$ is processed by a sinusoidal position embedding, producing $t_{emb} \in \mathbb{R}^{C_{emb}}$, which is repeated across the size $(w_r, h_r)$ and concatenated with $\mathbf{I}_{r}$ and $\mathbf{I}_{c}$, resulting in a feature of $\mathbb{R}^{w_r \times h_r\times(C_{emb}+6)}$.
After passing through a convolutional neural network (CNN), the feature is used to estimate the robot's 2D displacement in pixel space, $\mathbf{v}_{xy}\in\mathbb{R}^{2}$, rotation $\Delta\mathbf{R}_{:6}\in\mathbb{R}^{6}$, and depth of $\mathbf{v}_z\in\mathbb{R}$.
An update strategy for $\mathbf{H}_t=(\mathbf{R}_{t},\mathbf{t}_{t})$ to $\hat{\mathbf{H}}_{0}=(\hat{\mathbf{R}}_{0},\hat{\mathbf{t}}_{0})$ is as follows:
\begin{equation}
\begin{aligned}
    & \hat{\mathbf{t}}^{xy}_0 = (\frac{\mathbf{v}_{xy}}{f} + \frac{\mathbf{t}^{xy}_t}{\mathbf{t}^{z}_t} ) \hat{\mathbf{t}}^{z}_0 \\
    & \hat{\mathbf{R}}_0 = \Delta\mathbf{R}\mathbf{R}_t, \hat{\mathbf{t}}^{z}_0=\mathbf{v}_z\mathbf{t}^{z}_t,
    \label{eq:pose_denoising}
\end{aligned}
\end{equation}
where $\Delta\mathbf{R}$ is the rotation matrix obtained by Gram-Schmidt orthogonalization of the prediction $\Delta\mathbf{R}_{:6}$, as in Eq.~\ref{monocular-denormalized-rotation}, and $\mathbf{v}_z$ follows an incremental refinement of depth.
\newline
\textbf{Loss function.}
The pose denoising network is trained to predict the pose $\mathbf{H}_{0}$ rather than noise, which is a variational diffusion model with a target prediction type, following \cite{jiang2023se3}.
The pose loss is based on the distance between transformed points, defined as $\mathrm{Dist}(\mathbf{T}_1, \mathbf{T}_2) = \frac{1}{|\mathcal{X}|} \sum_{x \in \mathcal{X}} |\mathbf{T}_1 x - \mathbf{T}_2 x|$, where $\mathcal{X}$ is a set of  predefined points sampled from robot 3D model.
The final pose prediction loss function consists of specialized losses for each prediction $\mathbf{v}_{xy}$, $\Delta\mathbf{R}_{:6}$ and $\mathbf{v}_z$:
\begin{equation}
\begin{aligned}
    \mathcal{L} = \mathrm{Dist}(\mathbf{H}_0, \hat{\mathbf{H}}^{\mathbf{v}_{xy}}_0) + \mathrm{Dist}(\mathbf{H}_0, \hat{\mathbf{H}}^{\Delta\mathbf{R}_{:6}}_0) +\mathrm{Dist}(\mathbf{H}_0, \hat{\mathbf{H}}^{\mathbf{v}_z}_0).
    \label{eq:full_supervised}
\end{aligned}
\end{equation} 
Note that $\hat{\mathbf{H}}^{\mathbf{v}_{xy}}_0$ is the updated pose using Eq.~\ref{eq:pose_denoising}, based only on the predicted $\mathbf{v}_{xy}$, along with the ground-truth $\Delta\mathbf{R}_{:6}$ and $\mathbf{v}_z$, and similarly for $\hat{\mathbf{H}}^{\Delta\mathbf{R}_{:6}}_0$ and $\hat{\mathbf{H}}^{\mathbf{v}_z}_0$.

\section{Experiments}
\label{sec:experiment}

\subsection{Implementation details}
Our diffusion process uses a 100-step linear noise schedule with parameters $\mathrm{c}_z=1.5$ and $\gamma=3$. 
%
The pose denoising network uses a sinusoidal time encoder ($C_{emb}=4$), and ResNet34 CNN backbone.
%
The rendered image size is 320$\times$240, and the input image size is 640$\times$480.
We train the model for 300 epochs with a learning rate of $3\times10^{-4}$.
During inference, we employ reverse diffusion with DDIM sampling over 5 steps at $\eta=1.0$ to obtain a coarse estimation, followed by an additional 5 refinement steps of direct regression for fine-grained pose estimation.
Following Robopose~\cite{robopose}, we use 10 total iterations and identical pose initialization for fair comparison.
%
%
All code is implemented in PyTorch and executed on four NVIDIA RTX TITAN GPUs.

\begin{table*}[]
\caption{Evaluation results on real-world datasets.}
\hspace*{1pt} 
  \begin{minipage}{\linewidth} 
  
    \centering
    \setlength{\tabcolsep}{6pt} 
    \small
    \begin{tabular}{l@{\hskip 3pt}|c@{\hskip 6pt}c@{\hskip 5pt}c@{\hskip 5pt}c|cccccc}
    \toprule
            & \raisebox{-1ex}{Known} & \raisebox{-1ex}{Bound-} & \raisebox{-1ex}{Robot} & \raisebox{-1ex}{DREAM-real}              & \multicolumn{2}{c}{RealSense-Franka} & \multicolumn{2}{c}{AzureKinect-Franka} & \multicolumn{2}{c}{DREAM-real}    \\ 
    \vspace{-7pt} & & &  &        & \multicolumn{2}{c}{\rule[6pt]{2.5cm}{0.05pt}} & \multicolumn{2}{c}{\rule[6pt]{2.5cm}{0.05pt}} & \multicolumn{2}{c}{\rule[6pt]{2.5cm}{0.05pt}}  \\
    Method  & \raisebox{1ex}{joints} & \raisebox{1ex}{ing box} & \raisebox{1ex}{mesh} & \raisebox{1ex}{training}              & AUC \textcolor{green}{$\uparrow$}  & Mean(m) \textcolor{red}{$\downarrow$}  & AUC \textcolor{green}{$\uparrow$}  & Mean(m) \textcolor{red}{$\downarrow$}  & AUC \textcolor{green}{$\uparrow$}  & Mean(m) \textcolor{red}{$\downarrow$}   \\ \midrule
    HoliPose & \ding{55} & \ding{51} & \ding{55} &  \cellcolor{red!20}\ding{51}              & -            & -             & -              & -             & 76.18          & 0.024           \\
    RoboPEPP & \ding{55} & \ding{51} & \ding{55} &  \cellcolor{red!20}\ding{51}              & -            & -             & -              & -             & 77.67          & 0.026           \\
    RoboPose & \ding{55} & \ding{55}   & \ding{51}  &   \cellcolor{green!20}\ding{55}            & 37.93            & 0.121             & 39.28              & 0.135             & 71.58         & 0.029            \\
      \midrule
    CtRNet & \ding{51} & \ding{55} & \ding{51} & \cellcolor{red!20}\ding{51}           & 43.09            & 0.089             & 25.71              & 0.104             & 85.96          & 0.020          \\
    CtRNet-X & \ding{51} & \ding{55} & \ding{51} & \cellcolor{red!20}\ding{51}         & 48.49            & 0.066             & 35.83              & 0.125             & 86.23          & 0.014          \\ 
    HoliPose & \ding{51} & \ding{51} & \ding{55} & \cellcolor{red!20}\ding{51}         & 47.79            & 0.056             & 25.91              & 0.087             & 87.81          & 0.015       \\
    HoliPose & \ding{51} & \ding{51} & \ding{55} &  \cellcolor{green!20}\ding{55}              & 55.75            & 0.046             & 49.00              & 0.052             & 55.00          & 0.048          \\
    DREAM-Q & \ding{51} & \ding{55} & \ding{55} & \cellcolor{green!20}\ding{55}              & 0.0              & Inf               & 0.0                & Inf               & 56.98          & 59.28          \\
    DREAM-F & \ding{51} & \ding{55} & \ding{55} & \cellcolor{green!20}\ding{55}                & 0.15             & Inf               & 1.27               & Inf               & 60.74          & 113.0          \\
    DREAM-H & \ding{51} & \ding{55} & \ding{55} & \cellcolor{green!20}\ding{55}                & 9.839            & 852.8             & 3.478              & 1295              & 68.58          & 17.47          \\
    SGTAPose & \ding{51} & \ding{55} & \ding{55} & \cellcolor{green!20}\ding{55}               & 32.72            & Inf               & 25.30              & Inf               & 66.52          & 0.045          \\
    RoboPose & \ding{51} & \ding{55} & \ding{51} &   \cellcolor{green!20}\ding{55}            & 59.97            & 0.077             & 50.47              & 0.187             & 80.13         & 0.020            \\
    \rowcolor{lightgray}
    Ours     & \ding{51} & \ding{55} & \ding{51} &   \ding{55}            & \textbf{69.03}   & \textbf{0.0317}   & \textbf{66.75}     & \textbf{0.033}    & \textbf{81.18} & \textbf{0.018}  \\ 
    \bottomrule
    \end{tabular}
    \label{tab:main_result}
  \end{minipage}
\newline \newline
\centering
\end{table*}

\subsection{Datasets and metrics}
\label{sec:datasets_evaluation}
\textbf{Training dataset.}
%
We use a synthetic dataset (100k images) generated by DREAM~\cite{DREAM}, covering diverse robot poses and joint configurations.
%
%
While RoboKeyGen~\cite{robokeygen} introduces its own synthetic set, we use the DREAM-synt dataset to ensure fair comparison for all methods.
\newline
\textbf{Evaluation datasets.} 
We evaluate using two benchmarks (DREAM~\cite{DREAM}, RoboKeyGen~\cite{robokeygen}) with annotations for real-world Franka Panda robot.
%
%
%
%
%
%
%
The RoboKeyGen benchmark contains RealSense-Franka and AzureKinect-Franka datasets, presenting greater challenges as the robots exhibit low visual saliency against the background. 
%
\newline
\textbf{Metric.}
We evaluate pose estimation accuracy using the 3D reconstruction ADD metric, which computes distances between ground truth and predicted 3D keypoints at robot joint positions.
%
For comprehensive evaluation, we adopt both the AUC of ADD scores across 0.01\,--\,100~mm thresholds and the mean ADD value to assess cumulative and absolute pose accuracy, respectively.
\begin{figure}[t]
  \centering
   \includegraphics[width=\linewidth]{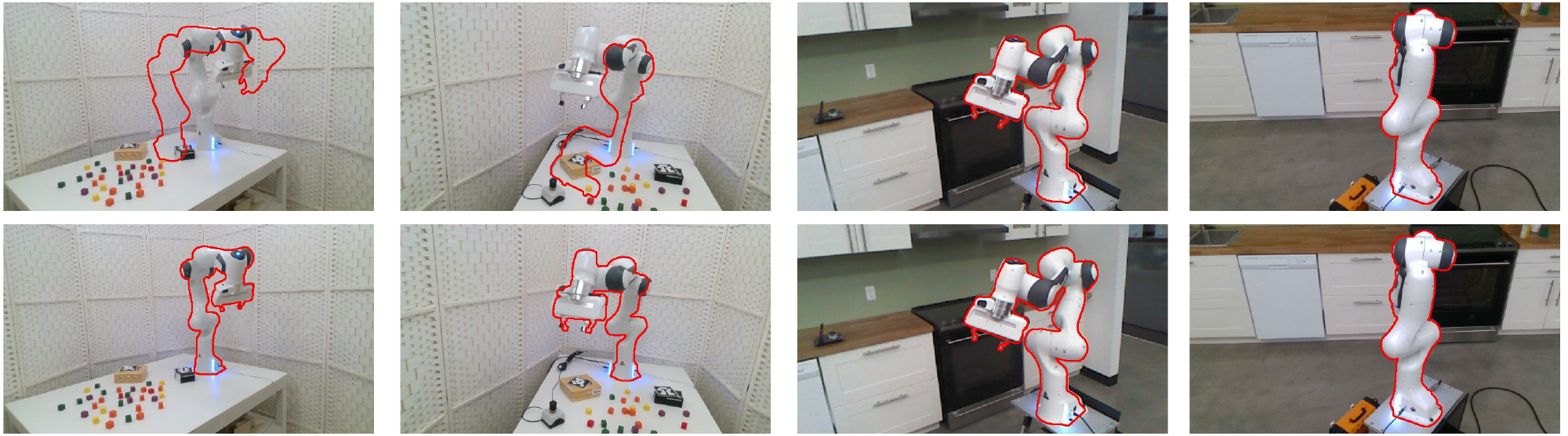}
   \vspace{-0.15cm}
   \highlight{
   {\footnotesize (a) Pose estimation results from RoboPose (top) and our method (bottom) on the RoboKeyGen (left) and DREAM (right) datasets. } 
   }
   \includegraphics[width=\linewidth]{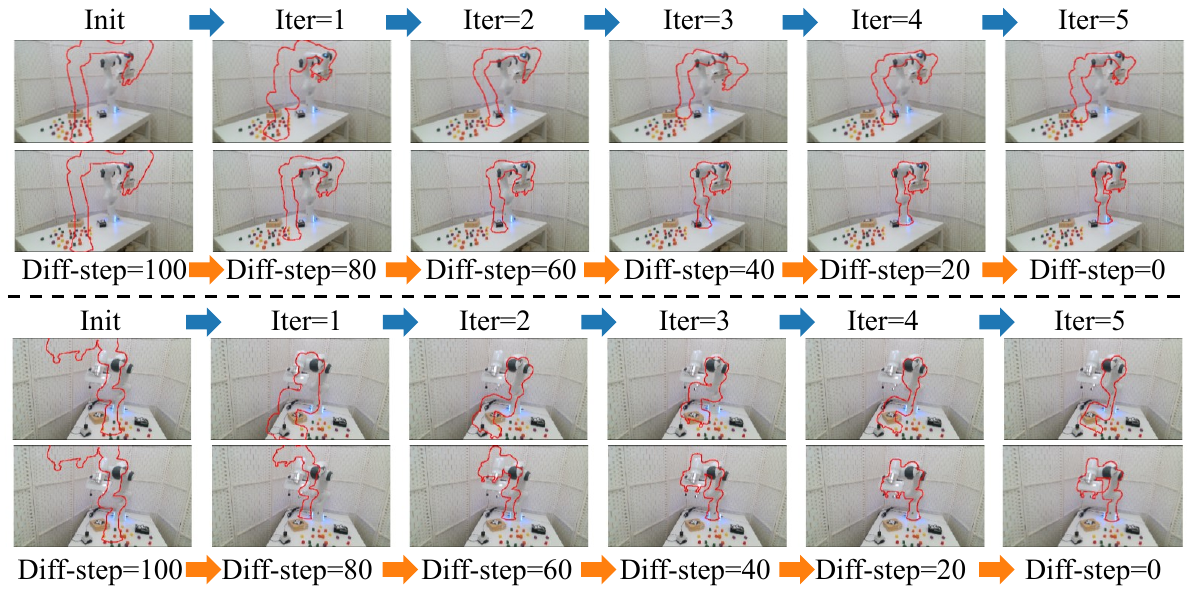}
   \vspace{-0.15cm}
   \highlight{
   {\footnotesize (b) 
    Pose estimation processes of two samples, for \textcolor[HTML]{1f77b4}{RoboPose} performing direct refinement and \textcolor[HTML]{ff7f0e}{Ours} using reverse diffusion.
   }
   }
   \vspace{-0.15cm}
    \highlight{\caption{Visualization of pose estimation results and processes.}}
   \vspace{-0.5cm}
   \label{fig:viualization}
\end{figure}

\subsection{Main experiment}
Tab.~\ref{tab:main_result} compares our approach with baselines across all real-world datasets, including their input assumptions.
We use the original authors' provided weights and inference parameters for all methods.
\newline
\textbf{Note}: 
%
%
(\textbf{i}) RoboKeyGen~\cite{robokeygen} (2024 ICRA) was not included as a baseline due to differences in its training dataset.
(\textbf{ii})``Inf'' represents an infinite distance between the prediction and the ground truth, which indicates a failed pose estimation.
\newline
\textbf{Results.}
Our method achieves state-of-the-art performance on all real-world datasets, with particularly significant gains on the RealSense-Franka and AzureKinect-Franka datasets.
%
%
On the most challenging AzureKinect dataset, our method achieves an AUC of 66.75, representing a 16.28-point (32.3\%) improvement over the best baseline.
%
%
%
These results demonstrate our method's robustness and accuracy across diverse datasets, with particularly strong performance on challenging cases.
%
Pose estimation comparisons are presented in Fig.~\ref{fig:viualization}, with detailed process illustrated in the Appendix.
%

The experimental results not only validate our method's superiority but also yield two important findings:
(\textbf{i}) 
RoboKeyGen datasets exhibit substantially greater challenge than DREAM, as evidenced by consistent performance degradation across all methods.
%
(\textbf{ii}) 
While self-supervised training on DREAM-real enhances performance on seen data, it compromises generalization capability on unseen data, as demonstrated by HoliPose's decreasing performance on RoboKeyGen.
Thus, we do not use self-supervised learning in our method.

\begin{figure}[t]
  \centering
   \includegraphics[width=\linewidth]{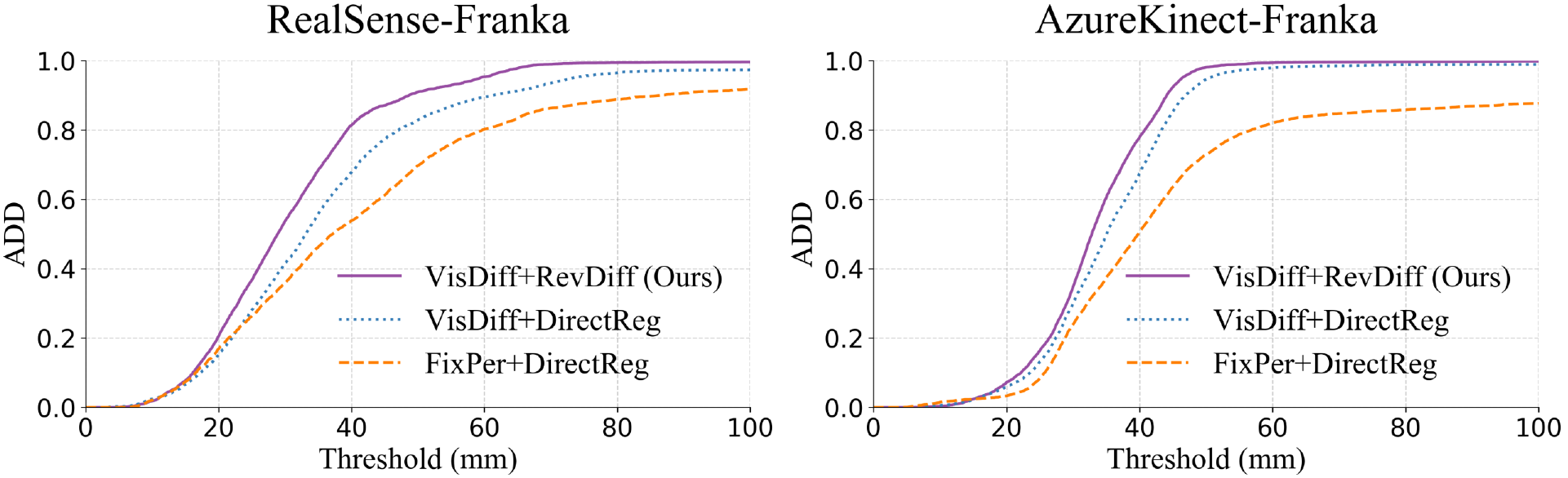}
    \caption{Distribution of ADD scores on RoboKeyGen benchmark.}
   \vspace{-0.3cm}
   \label{fig:ablation_study}
\end{figure}

\begin{table}
    \centering
    \caption{AUC results across ablation variants.}
    {\fontsize{9}{11}\selectfont
    \begin{tabular}{cc|cc|cc}
    \toprule
       VisDiff  & FixPer    & RevDiff & DirectReg    & Real  & Azure \\
    \midrule
       \ding{55}  & \ding{51} & \ding{55} & \ding{51}  & 56.96 & 53.44 \\
       \ding{51}  & \ding{55} & \ding{55} & \ding{51}  & 63.53 & 64.18 \\
       \ding{51}  & \ding{55} & \ding{51} & \ding{55}  & 69.03 & 66.75 \\
    \bottomrule
    \end{tabular}
    }
    \vspace{-0.5cm}
    \label{tab:ablation_study}
\end{table}

\subsection{Ablation study}
Our approach achieves substantial improvements over baselines, benefiting from two key advantages:
(\textbf{i}) 
The visibility-constrained diffusion process (VisDiff in Sec.~\ref{sec:method-2}) generates high quality training poses and significantly enhances the model generalization.
(\textbf{ii}) 
The timestep-aware reverse diffusion process (RevDiff in Sec.~\ref{sec:method-3}) enables scheduled, coarse-to-fine pose estimation and effectively improves prediction robustness. 

To validate advantage (\textbf{i}), we employ a fixed-scale pose perturbation (FixPer) introduced by RoboPose~\cite{robopose}, which samples a noisy transformation from a fixed-scale normal distribution to perturb the ground truth at the training time.
To validate advantage (\textbf{ii}), we employ 10 times direct regression (DirectReg) to refine the pose.
%
%
All other training settings remain identical to ensure fair comparison.
%
The overall distributions of ADD and AUC results on the RoboKeyGen benchmark are presented in Fig.~\ref{fig:ablation_study} and Tab.~\ref{tab:ablation_study}, respectively, with RealSense-Franka and AzureKinect-Franka abbreviated as \textit{Real} and \textit{Azure}.
%
We find that VisDiff substantially improves the accuracy, while RevDiff further enhances pose estimation performance.


\begin{table}
    \centering
    \caption{AUC and inference speed of different steps iteration.}
    {\fontsize{9}{11}\selectfont
        \begin{tabular}{cc|ccc}
        \toprule
        Method & Num of Steps & Real & Azure & Speed(FPS) \\
        \midrule
        RoboPose&10        &  59.97 & 50.47    & 0.67  \\
        Ours &10           &  69.03  & 66.75    & 0.65           \\
        Ours &5            &  66.65 & 65.74    & 1.30           \\
        Ours &1 (tracking) &  69.38 & 67.04    & 6.50           \\
        \bottomrule
        \end{tabular}
    }
    \label{tab:compution_cost}
    \vspace{-0.3cm}
\end{table}

\subsection{Analysis of computational cost}
We evaluated the AUC and average runtime (FPS) on the RoboKenGen benchmark using our hardware configuration: an Intel i5-2.2GHz CPU paired with a TITAN RTX GPU, as shown in Tab.~\ref{tab:compution_cost}.
Benefiting from DDIM sampling (Eq.~\ref{eq:monocular-Euclidean-reverse}), we can control the number of denoising steps
%
to 5 and achieves 2× faster inference with $\Delta$AUCs -2.38 and -1.01 in two datasets. 
%
Following the diffusion-based motion planner~\cite{diffuser-planner}, in scenarios where the robot pose is continuously tracked from streamed image observations,  we can initialize the pose using the previous estimate and apply a single-step denoising, with $\Delta$AUCs +0.35 and +0.29.
%
%
The speeds of other baselines are as follows: DREAM -- 5.2~FPS, HoliPose -- 7.3~FPS.
%
Our reported speeds are lower than HoliPose's due to GPU differences (TITAN RTX vs. V100S).

\section{Conclusion}
\label{sec:conclusion}
In this work, we proposed MonoSE(3)-Diffusion, a monocular SE(3) diffusion framework for robust camera-to-robot pose estimation, which is formulated as a conditional denoising diffusion process in the viewing frustum.
%
This framework introduces a visibility-constrained diffusion process, ensuring training poses were diverse and distributed within the frustum for comprehensive model training.
To constrain poses in the visible view, we introduced the monocular-normalization associated with SE(3) for the diffusion process.
%
During the inference phase, our timestep-aware reverse diffusion process demonstrated robust pose estimations, guided by the conditional pose distribution at each diffusion step.
%
The pose prediction of a rendering-based denoiser network was also facilitated by the timestep condition.
%
%
%
Our method outperforms existing approaches, achieving 15.1\% and 32.3\%  improvements on two datasets from the challenging RoboKeyGen benchmark.

\begin{figure}[h]
      \centering
       \includegraphics[width=\linewidth]{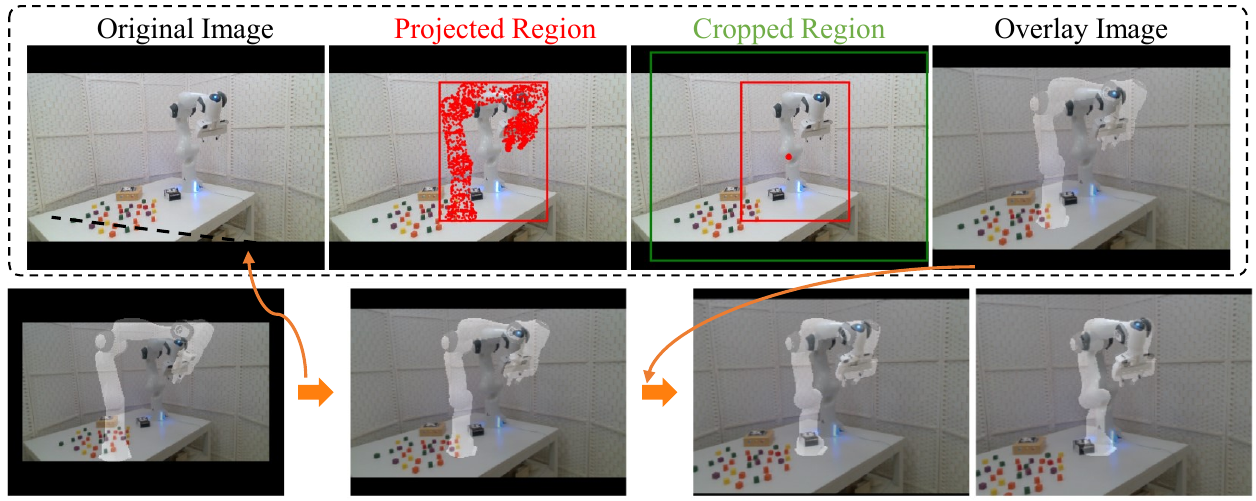}
       \caption{
        \textbf{Top}: Illustration of a single iteration in the reverse diffusion, including: (1) the original input image, (2) projected 3D points and the resulting red bounding box (projected region), (3) corresponding cropped green box centered on the robot projection, and (4) an overlay between the cropped image and the rendering.
        \textbf{Bottom}: Overlays between the cropped image and the rendered image across successive denoising steps. 
         }
       \label{fig:crop}
       \vspace{-0.3cm}
    \end{figure}

\appendix
As shown in Fig.~\ref{fig:crop}, we present a qualitative example that illustrates the denoising process of reverse diffusion-based pose estimation.
%
%
%
    %
    %
    %
    %
    
        %

%







\ifCLASSOPTIONcaptionsoff
  \newpage
\fi



%


\bibliographystyle{IEEEtran}
\bibliography{myref}

%








\end{document}